\documentclass{ecai} 

\usepackage{graphicx}
\usepackage{latexsym}
\usepackage{amssymb}
\usepackage{amsmath}
\usepackage{amsthm}
\usepackage{booktabs}
\usepackage{enumitem}
\usepackage{color}
\usepackage{multicol}
\usepackage{multirow}
\usepackage{booktabs}

\usepackage{times}
\usepackage{latexsym}
\usepackage[T1]{fontenc}
\usepackage[utf8]{inputenc}
\usepackage{microtype}
\usepackage{inconsolata}
\usepackage[colorinlistoftodos,prependcaption,textsize=tiny]{todonotes}
\usepackage{tcolorbox}
\usepackage{pifont}
\usepackage{xcolor}
\usepackage{hhline}
\usepackage{regexpatch}
\usepackage{amssymb}
\usepackage{amsmath}
\usepackage{multirow}
\usepackage{booktabs}
\makeatletter
\xpatchcmd{\@todo}{\setkeys{todonotes}{#1}}{\setkeys{todonotes}{inline,#1}}{}{}
\makeatother
\usepackage{footnote}
\usepackage{chngcntr}
\usepackage{adjustbox}
\usepackage{float}

\usepackage{svg}

\usepackage{caption}
\newcommand{\BibTeX}{B\kern-.05em{\sc i\kern-.025em b}\kern-.08em\TeX}

\begin{document}


\begin{frontmatter}


\paperid{2277} 


\title{Unpacking Robustness in Inflectional Languages: Adversarial Evaluation and Mechanistic Insights}


\author[A]{\fnms{Paweł}~\snm{Walkowiak}\orcid{0009-0008-0381-9202}\thanks{Corresponding Author. Email: pawel.walkowiak@pwr.edu.pl}}
\author[A]{\fnms{Marek}~\snm{Klonowski}\orcid{0000-0002-3141-8712}}
\author[A]{\fnms{Marcin}~\snm{Oleksy}\orcid{0000-0001-7740-5557}}
\author[A]{\fnms{Arkadiusz}~\snm{Janz}\orcid{0000-0002-9203-5520}} 

\address[A]{Wrocław University of Technology}


\begin{abstract}
Various techniques are used in the generation of adversarial examples, including methods such as TextBugger which introduce minor, hardly visible perturbations to words leading to changes in model behaviour. Another class of techniques involves substituting words with their synonyms in a way that preserves the text's meaning but alters its predicted class, with TextFooler being a prominent example of such attacks. Most adversarial example generation methods are developed and evaluated primarily on non-inflectional languages, typically English. In this work, we evaluate and explain how adversarial attacks perform in inflectional languages. To explain the impact of inflection on model behaviour and its robustness under attack, we designed a novel protocol inspired by mechanistic interpretability, based on Edge Attribution Patching (EAP) method. The proposed evaluation protocol relies on parallel task-specific corpora that include both inflected and syncretic variants of texts in two languages -- Polish and English. To analyse the models and explain the relationship between inflection and adversarial robustness, we create a new benchmark based on task-oriented dataset MultiEmo, enabling the identification of mechanistic inflection-related elements of circuits within the model and analyse their behaviour under attack.
\end{abstract}

\end{frontmatter}


\section{Introduction}
Transformer-based language models~\cite{DBLP:conf/nips/VaswaniSPUJGKP17} represent the state of the art in text classification. These models demonstrate exceptional performance on a wide range of natural language processing tasks. However, their robustness can be compromised by Adversarial Examples~\cite{DBLP:journals/corr/SzegedyZSBEGF13}. 
Carefully crafted adversarial examples for tasks such as classification can mislead the model and fool it to the point where it is not useful. Although numerous studies have explored methods for generating and defending against adversarial examples, their behavior in inflectional languages remains unexplored. Another uninvestigated path is the method of accessing high-quality adversarial examples, researchers tend to use different ways of filtering the candidates to the point that they are difficult to distinguish from the original. Currently used metrics are rely on various text meaning levels, including n-gram statistics~\cite{lin-2004-rouge}, semantics~\cite{reimers-2019-sentence-bert} and natural language inference~\cite{camburu2018esnli}. The results produced by these metrics may vary, highlighting the importance of understanding their respective strengths and limitations.


When selecting language models for specific tasks, decisions are often primarily guided by performance metrics such as accuracy. However, model robustness is frequently overlooked in this process. Among the wide range of models available, both language-specific models~\cite{DBLP:journals/corr/abs-1810-04805, mroczkowski-etal-2021-herbert, DBLP:conf/tsd/StrakaNSS21_Robeczech} and multilingual models~\cite{DBLP:journals/corr/abs-1911-02116} can be considered. This raises the question of whether particular characteristics of a model's training language, such as inflectional complexity, might influence its robustness to text-based adversarial attacks.  In this work, our aim is to investigate this question by designing adversarial examples~\cite{DBLP:journals/corr/SzegedyZSBEGF13} and gaining insights from the emerging field of model explainability, Mechanistic Interpretability~\cite{DBLP:journals/corr/abs-2410-09087}. 


\begin{figure}[t!]
\centering
\includegraphics[width=0.9\linewidth, height=70mm]{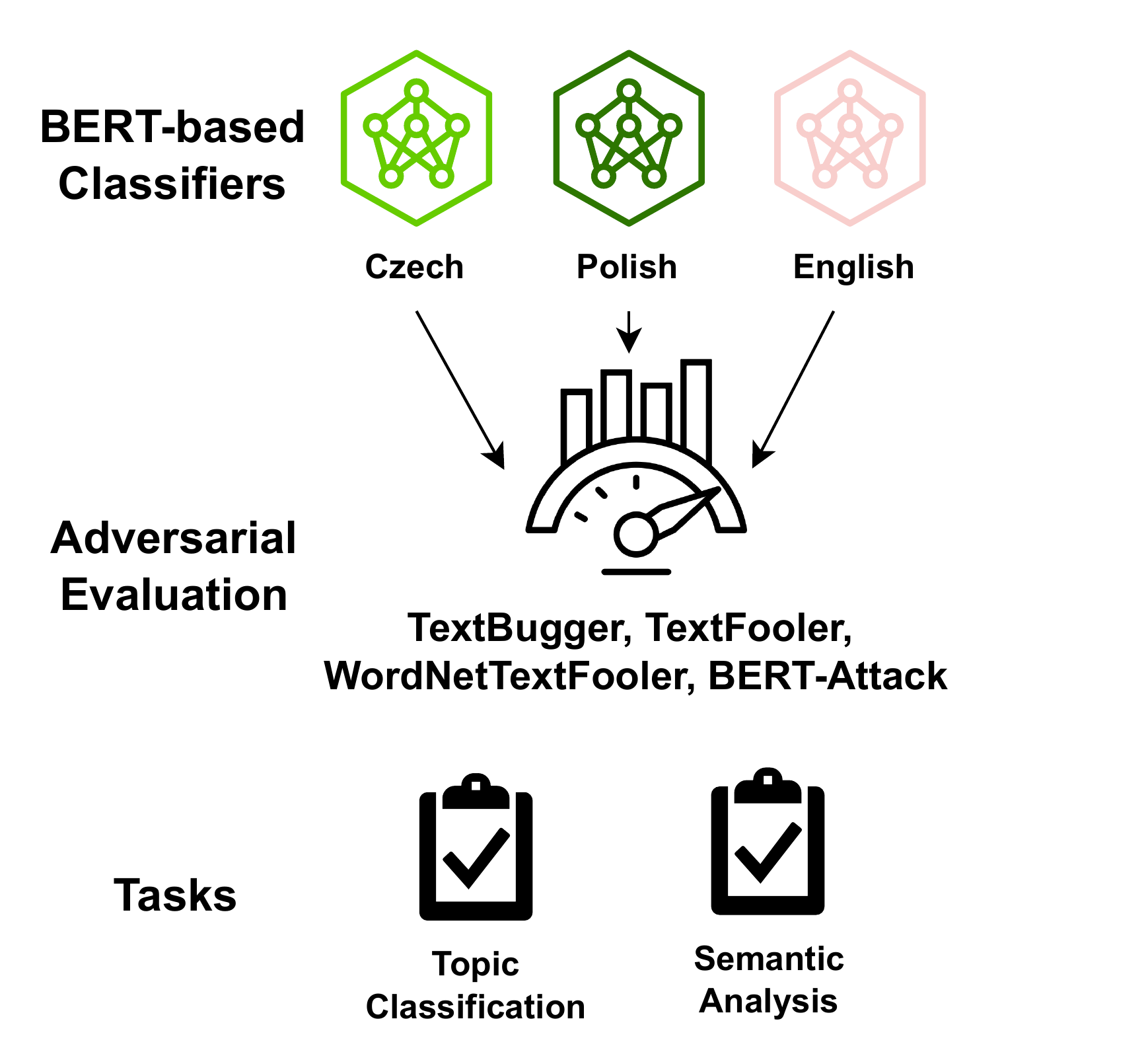}
\caption{Experimental pipeline for evaluation of adversarial examples generation methods.} \label{idea}
\end{figure}

To answer our research question, we evaluated the performance of various text attack methods~\cite{DBLP:journals/corr/abs-1812-05271, DBLP:conf/emnlp/LiMGXQ20, DBLP:journals/corr/abs-1907-11932, DBLP:conf/ecai/GniewkowskiSKW23} across three selected languages: two inflectional languages (Polish and Czech) and English, which serves as a widely used, non-inflectional reference language. Using techniques from Mechanistic Interpretability, particularly the circuit discovery approach EAP-IG~\cite{hanna2024have} inspired by the work of Olah et al.~\cite{olah2020zoom}, we identified components within the multilingual model XLM-RoBERTa~\cite{DBLP:journals/corr/abs-1911-02116} that are responsible for processing inflectional information in textual inputs. Building on this analysis, we investigate how circuits incorporating inflection-related elements compare to those without them in terms of adversarial robustness. In this paper, we present a comprehensive set of experiments aimed at understanding how the presence of inflectional morphology in a language influences the adversarial robustness of language models. By comparing models’ responses to adversarial attacks in both inflectional (e.g. Polish and Czech) and non-inflectional (e.g. English) languages, we seek to uncover language-specific vulnerabilities and mechanisms within the model. Additionally, we employ tools from the emerging field of Mechanistic Interpretability to identify the internal structures responsible for processing inflectional features. Our key contributions are as follows:

\begin{itemize}
    \item \textbf{A comprehensive evaluation and comparison of four adversarial example generation methods} (TextFooler, TextBugger, WordNetTextFooler and BERT-Attack) across two cross-lingual NLP tasks, with a focus on topic classification and sentiment analysis in both highly inflectional (Polish and Czech) and non-inflectional (English) languages.
    \item Evaluation of widely used similarity assessment methods: \textbf{ROUGE, Semantic Similarity and Natural Language Inference} for their effectiveness in measuring the quality of adversarial examples.
    \item We introduce \textbf{a novel parallel dataset specifically designed to facilitate the mechanistic identification of inflectional components} within task-specific language model circuits, enabling fine-grained analysis of model robustness.
    \item \textbf{Mechanistic interpretation of inflection's impact on model robustness} using Edge Attribution Patching (EAP) technique and our novel dataset.

\end{itemize}

\section{Related Work}

\paragraph{Text Classification}
The classification task involves assigning input samples $X = \{X_1, X_2, ..., X_m\}$ to one of the $n$ classes $Y = \{Y_1, Y_2, ... , Y_n\}$, or to multiple classes in the case of multi-label classification. A classifier performs a mapping $X \rightarrow Y$ aiming to generalize to previously unseen data, based on the knowledge acquired during the training process on a labeled training corpus. Classification plays a crucial role in many tasks of natural language processing (NLP), such as sentiment analysis~\cite{DBLP:journals/corr/TaiSM15} and topic classification~\cite{DBLP:journals/isci/ChenGL19}. The introduction of the Transformer model in~\cite{DBLP:conf/nips/VaswaniSPUJGKP17} marked a significant advance in the development of high-quality text classification systems. Building on this foundation, the BERT architecture~\cite{DBLP:journals/corr/abs-1810-04805} was introduced, which achieves state-of-the-art performance in a wide range of NLP tasks. BERT is particularly effective for text classification because of its pretraining on large-scale text corpora, which enables it to learn general language representations. These representations can then be fine-tuned for specific tasks such as text classification, resulting in high accuracy even when only limited amounts of labeled data are available, which is a valuable advantage given the cost and effort required for manual annotation.

Recently, decoder-only transformer models such as GPT-3.5~\cite{openai2023gpt35} have demonstrated increasing performance across a variety of NLP tasks. This has led some to question whether BERT-based models remain a state-of-the-art solution for text classification. However, studies such as~\cite{DBLP:conf/emnlp/LukitoCMS24, DBLP:journals/corr/abs-2411-05050} have shown that BERT-based models can still outperform decoder-only architectures in text classification tasks. These findings underscore the continued relevance of evaluating the robustness of BERT-based approaches, given their ongoing use in both academic research and practical applications.


\paragraph{Adversarial Examples in Text Classification}
Adversarial examples (AEs) are input samples that have been subtly perturbed in a way that is often imperceptible to humans but causes a model to produce incorrect predictions, deviating from the original correct output. They are currently considered a critical threat to the performance of the models and their trustworthiness. Since the influential work presented in~\cite{szegedy2014intriguingpropertiesneuralnetworks}, the field has seen the publication of tens of thousands of related studies. A substantial number of these contributions are catalogued in the continuously expanding list provided in~\cite{carlini}. Most adversarial example generation methods exploit vulnerabilities in model architectures by introducing subtle perturbations into the input text to induce misclassification. Among the growing number of adversarial examples generation methods, we focus on four of them specifically designed to target BERT-based models: TextFooler~\cite{DBLP:journals/corr/abs-1907-11932}, TextBugger~\cite{DBLP:journals/corr/abs-1812-05271}, WordNetTextFooler~\cite{DBLP:conf/ecai/GniewkowskiSKW23}, and BERT-Attack~\cite{DBLP:conf/emnlp/LiMGXQ20}.



\paragraph{Inflection} In the context of natural language processing, inflection refers to the variation in the form of a word to express grammatical features such as tense, number, gender, case, or person. These changes typically occur through the addition of affixes (such as suffixes) or by internal modification of the word. From a linguistic perspective, inflection is the variation of words used to indicate relationships between them in a sentence through the application of a set of word-forms, suffixes, and prefixes. 
Inflection is an important aspect of natural language processing, because it affects tasks such as part-of-speech tagging, syntactic parsing, information retrieval, and machine translation. A key feature of inflectional languages in text generation is that a single morpheme (the smallest unit of meaning) can take on multiple functions through different inflectional endings, making accurate modification crucial to preserve sentence meaning during attacks. Additionally, the relative freedom in word order allows for paraphrasing while maintaining semantic content. These characteristics increase syntactic complexity and require more sophisticated approaches for attacks generation.



 

\paragraph{Mechanistic Interpretability}
Mechanistic Interpretability (MI) is a subfield of Artificial Intelligence Explainability (XAI) that aims to understand how AI models work at a detailed mechanistic level. The goal is not just to observe what inputs lead to what outputs, but to uncover the internal processes and structures (e.g., neurons, weights, attention heads, and MLPs) that give rise to a model's behavior. The definition of MI and what can be called the mechanistic method was examined in~\cite{DBLP:journals/corr/abs-2410-09087}. One of the central concepts in mechanistic interpretability, as introduced in~\cite{olah2020zoom} is the notion of a \textbf{circuit}. A circuit refers to a specific subset of components of the neural network that collectively execute a narrow, well-defined task. An isolated subset of model components is generally easier to interpret, as it allows a clearer analysis of how these elements interact to contribute to the model’s output. One of the earliest methods for identifying task circuits was ACDC~\cite{DBLP:conf/nips/ConmyMLHG23}. More recently, the work of~\citep{nanda2023} introduced attribution patching in a method called \textbf{Edge Attribution Patching (EAP}), an efficient approximation of activation patching that significantly reduces the computational cost associated with circuit discovery.


\section{Adversarial Examples Generation Methods}
\paragraph{TextFooler} (TF) introduced in \cite{DBLP:journals/corr/abs-1907-11932}, are methods to generate adversarial examples against text BERT-based classifiers. TextFooler identifies the most important words in a text-based model decision change on masked words. For each word, iteration is found in decreasing order of importance, defined as the word's impact on the model's decision marked as correct synonym using uncontextual embedding from the Glove model~\cite{pennington-etal-2014-glove}. Adversarial samples are generated by replacing words with their substitutes in the original sentence.  

\paragraph{TextBugger} (TB) is an adversarial example generation method introduced in \cite{DBLP:journals/corr/abs-1812-05271}. The candidate word for attack is selected with the usage of Jacobian $J_F(i, y) = \frac{\partial F_y(X)}{\partial x_i},$ where $X = (x_1, x_2, ... x_n)$ is input text, which consists of words $\{x_1, x_2, ... x_n\}$ , $F(X)$ is classification function and $y$ model prediction. Based on the ranking, perturbations are introduced into important words in order to prepare an attack. The TextBugger proposes five perturbation methods: \textit{insertion}, which involves adding a space within the perturbed word; \textit{deletion}, which removes a random character; \textit{swap}, which exchanges the positions of two adjacent characters; \textit{Sub-C}, which substitutes characters with visually similar ones (e.g., replacing "l" with "1"); and \textit{Sub-W}, which replaces a word with its nearest neighbour based on a GloVe model~\cite{pennington-etal-2014-glove}.  

\paragraph{WordNetTextFooler} (WNTF) an adversarial example generation method, proposed in \cite{DBLP:conf/ecai/GniewkowskiSKW23}. The method is based on the approaches introduced in TextFooler~\cite{DBLP:journals/corr/abs-1907-11932} and TextBugger~\cite{DBLP:journals/corr/abs-1812-05271}. However, it uses a different technique to generate word importance rankings and incorporates the aspect of inflectional variation of suggested synonyms, using Polish WordNet~\cite{wordnet}. In contrast to the importance score estimation techniques used in previous methods, the authors of WordNetTextFooler propose using a model explainability method Shapley values~\cite{lundberg2017unified} to assess the importance of words. These values are computed per model on a designated subset of the data. The method replaces the candidate words with the appropriate synonyms drawn from WordNet.

\paragraph{BERT-Attack} (BA) \cite{DBLP:conf/emnlp/LiMGXQ20} is an adversarial example generation method that takes advantage of a BERT-type model to generate possible word substitutions in a text. The approach is innovative in that it employs a pretrained language model to attack another language model, unlike earlier substitution methods that rely on simple perturbations or synonym replacements based solely on word vectors. The use of a contextualized language model allows for substitutions that take into account the surrounding text, yielding highly accurate replacements. The vulnerable words are selected by checking how the model prediction changes when masking each of the candidate words. The vulnerable words are replaced with candidate BERT generated tokens.

\section{Experiments}

The experiments are divided into two parts: 
\begin{itemize}
    \item Evaluating the effectiveness of adversarial attacks on models trained for the same tasks across selected languages, to compare robustness as influenced by inflectional morphology.
    \item Identifying the model components responsible for processing inflection and assessing their robustness to adversarial examples.
\end{itemize}

\subsection{Datasets}


In the first part of our experiments, we evaluated the effectiveness of adversarial example generation methods in reducing the robustness of BERT-based models. This evaluation was carried out on two natural language processing tasks: topic classification and semantic analysis. The Polish datasets were \textbf{Wiki\_PL}\footnote{\url{http://hdl.handle.net/11321/222}}, which consists of texts obtained from the Polish Wikipedia, which have been tagged in one of 34 thematic categories and \textbf{MultiEmo} the Polish part of the multilingual corpus \cite{multiemo}, used for text sentiment analysis. For the Czech language we choose \textbf{Czech Text Document Corpus v 2.0} \cite{DBLP:journals/corr/abs-1710-02365}, a data set containing annotated news articles from the Czech press, and The collection of Czech movie reviews from the csfd.cz portal, \textbf{CSFD}\footnote{\url{https://corpora.kiv.zcu.cz/sentiment/csfd.zip}}, which contains user reviews, along with sentiment annotations classified into one of three categories.
The classification on English texts were conducted on the \textbf{AG\_News}\footnote{\url{http://groups.di.unipi.it/~gulli/AG_corpus_of_news_articles.html}} and \textbf{IMDB} \cite{imdb} datasets. The AG\_News is a collection of news articles, labeled onto one of four thematic categories. The IMDB consists of short text movie reviews adapted to sentiment classification. To carry out the attacks, we trained classification models using pre-trained language models. Pretrained models used for the classification training where BERT\footnote{\url{https://huggingface.co/google-bert/bert-base-uncased}}~\cite{DBLP:journals/corr/abs-1810-04805} for English, HerBERT\footnote{\url{https://huggingface.co/allegro/herbert-base-cased}}~\cite{mroczkowski-etal-2021-herbert} for Polish and RobeCzech\footnote{\url{https://huggingface.co/ufal/robeczech-base}}~\cite{DBLP:conf/tsd/StrakaNSS21_Robeczech} for Czech.
The evaluation results of the base classification models, along with the subset sizes and characteristics of the dataset, can be found in Table \ref{tab:classification_results}. The datasets were divided into train and test parts. The test part was used in evaluation and generation of adversarial examples. 

The dataset employed for the mechanistic interpretability analysis of inflectional models is described in Section~\ref{icdd}.

\begin{table}[htb!]
\centering
\caption{The dataset characteristics include, the number of annotated classes, the sizes of subsets, the average word count, and the evaluation results of trained classifiers measured in terms of accuracy (ACC).\label{tab:classification_results}}
\begin{tabular}{|l|l|r|rr|r|r|}
\hline
\multicolumn{1}{|c|}{\multirow{2}{*}{Dataset}} & \multicolumn{1}{|c|}{\multirow{2}{*}{Lang.}}& \multicolumn{1}{c|}{\multirow{2}{*}{\begin{tabular}[c]{@{}c@{}} Class\\ number\end{tabular}}} & \multicolumn{2}{c|}{Dataset size}                                             &  \multicolumn{1}{c|}{\multirow{2}{*}{\begin{tabular}[c]{@{}c@{}} Avg.\\ words\end{tabular}}} & \multicolumn{1}{c|}{\multirow{2}{*}{ACC}} \\ 

\multicolumn{1}{|c|}{}  & \multicolumn{1}{c|}{} & \multicolumn{1}{c|}{} & \multicolumn{1}{c|}{train}   & \multicolumn{1}{c|}{test} & \multicolumn{1}{c|}{} & \multicolumn{1}{c|}{}                               \\ \hline

AG\_News & ENG & 4 & \multicolumn{1}{r|}{120,000} & \multicolumn{1}{r|}{4,788} & 38 & 94.72\\ \hline
IMDB & ENG &  2 & \multicolumn{1}{r|}{8,000}  & \multicolumn{1}{r|}{1,260} & 211 & 95,17 \\ \hline    
Wiki\_PL & PL &  34 & \multicolumn{1}{r|}{6,885} & \multicolumn{1}{r|}{1,859} & 186 & 94.88 \\ \hline
MultiEmo & PL &  2 & \multicolumn{1}{r|}{4,319}  & \multicolumn{1}{r|}{680} & 129 & 98.25\\ \hline
CTDC & CZ &  37 & \multicolumn{1}{r|}{11,955}   & \multicolumn{1}{r|}{1,574} & 247 & 83.33
\\ \hline
CSFD & CZ & 3 & \multicolumn{1}{r|}{82,242} & \multicolumn{1}{r|}{5,757} & 50 & 86.59 \\ \hline
\end{tabular}
\end{table}

\subsection{Adversarial Examples Performance}

In the experimental setup presented on Figure~\ref{idea}, the selected adversarial attack generation methods were implemented: \textbf{TextBugger}, \textbf{TextFooler}, \textbf{WordNetTextFooler} and \textbf{BERT-Attack}. We introduced a modification compared to the original methods, the unification of the identification of attack victim words. We used a masking algorithm that was applied to successive words in the texts, and changes in the confidence of the model, similar to the one proposed in the TextFooler method. For filtering adversarial example candidates to ensure their high quality, we used cosine similarity, with a threshold of 90\%, on text embeddings generated by the multilingual SBERT model\footnote{https://huggingface.co/sentence-transformers/distiluse-base-multilingual-cased-v1}.

The effectiveness of adversarial example generation methods was evaluated on a test subset of the considered datasets. We evaluated the degradation in classification quality -- measured in terms of classification accuracy -- by comparing the baseline models' performance on the original test set with their performance when exposed to adversarial examples. The results of accuracy change are presented in Table~\ref{tab:cls_change}. Baseline models exhibited accuracy levels that exceeded 94\% for models trained in Polish and English datasets and greater than 83\% for models trained on Czech datasets. \textbf{The greatest drop in classification models performance was caused by the TextBugger attack}. The results of the accuracy change correlate with the attack success rates obtained for the datasets. Similarly to the attack success rate, sentiment analysis in English was less robust to attacks. When comparing the results of language-specific models, we observe that in the topic classification task, the English model is more robust than the models for inflected languages such as Polish and Czech. However, in the semantic analysis task, the trend is reversed: the English model exhibits a significantly larger decrease in accuracy under attacks, whereas the Polish and Czech models demonstrate greater robustness. 

\begin{table}[tb]
\centering
\caption{Classification accuracy change after attacks, for each attack method.}
\label{tab:cls_change}

\begin{tabular}{|l|l|r|r|r|r|}
\hline
\multirow{2}{*}{Dataset} & \multirow{2}{*}{Lang.} & \multicolumn{4}{|c|}{$\Delta$ Accuracy {[}\%{]}} \\ \cline{3-6}
&& TB [\%] & TF [\%] & WNTF [\%] & BA[\%] \\ \hline

AG\_News & ENG & 1.90 & \textbf{8.44}  & 2.78 & 2.36  \\ \hline
IMDB & ENG  & 10.33 & \textbf{19.69} & 9.38 & 4.46  \\ \hline
Wiki\_PL & PL & 5.10 & \textbf{8.54}  & 5.10 & 2.03  \\ \hline
MultiEmo & PL  & 7.66 & \textbf{9.72}  & 4.13 & 2.81 \\ \hline
CSFD & CZ & 7.99 & \textbf{22.62} & 1.84 & 3.04  \\ \hline
CTDC & CZ & 5.38 & \textbf{10.84} & 0.86 & 4.23  \\ \hline
\end{tabular}
\end{table}

\subsection{How to Measure Adversarial
Examples Quality?}
One of the key conditions in generating adversarial examples is their high similarity to original text, in other words, corrupted text should maintain its meaning for the human reader, while inducing the model to alter its classification decisions. 
In this experiment, we want to address the issue of measuring adversarial examples' quality, by comparing three commonly used similarity measuring methods: \textbf{ROUGE}~\cite{lin-2004-rouge}, \textbf{Natural Language Inference (NLI)}~\cite{he2021deberta}, and \textbf{Semantic Similarity}~\cite{reimers-2019-sentence-bert}. The NLI score was based on the probability of entailment. For semantic similarity, we used cosine similarity based on embeddings generated by the SBERT model\footnote{https://huggingface.co/sentence-transformers/distiluse-base-multilingual-cased-v1}. The evaluation involved measuring the similarity between pairs of modified and unmodified texts, with results averaged for each attack method. The mean similarity scores for each language, along with their standard deviations, are presented in Table~\ref{table:sims}. The selected metrics aimed to analyse the modifications that led to changes in classifier decisions from different perspectives: ROUGE for statistical similarity, NLI for inference capability, and semantic similarity for meaning preservation.


More effective attacks, such as TextBugger, show lower statistical similarity values compared to less successful ones. In the case of similarity measured by confidence in the NLI classification, a noticeable difference emerges between languages, for English, the highest values are achieved by the TextBugger and WordNetTextFooler methods, while for Polish and Czech, TextBugger and BERT-Attack achieve the highest values. Furthermore, it is worth noting the high standard deviation for the NLI metric, suggesting significant variation between individual data samples (instability of the metric). For semantic similarity, the measured values are generally similar, with all values exceeding 96\%, and the standard deviation is low, within one and a half percentage points. An important observation from the experiment results is the high similarity of adversarial examples generated by the WordNetTextFooler method; these examples are more similar according to ROUGE and rank first or second in terms of semantic similarity, depending on the dataset. This observation suggests the high quality of the attacks generated by WordNetTextFooler, understood as the semantic proximity to the original examples.


\begin{table}[htb!]
    \caption{Similarity of adversarial examples to their originals, measured using four metrics. The results are presented by dataset and attack type. Values are calculated as the mean similarity of examples, with the standard deviation highlighted}
    \label{table:sims}
    \centering
    \begin{tabular}{|l|l|r|r|r|}
    \hline
    \multirow{2}{*}{DS} & \multirow{2}{*}{\begin{tabular}[c]{@{}c@{}}Attack\\type\end{tabular}} & \multicolumn{3}{|c|}{Similarity Metric} \\ \cline{3-5}
    
    & & ROUGE[\%] & NLI[\%] & Semantic[\%] \\ \hline

    \multirow{4}{*}{\rotatebox{90}{AG\_News}} 
    & TB & 87.81 $\pm$ 3.61 & \textbf{93.28} $\pm$ 13.35 & 96.51 $\pm$ 0.89 \\ 
    & TF & 92.81 $\pm$ 3.81 & 68.76 $\pm$ 34.58 & \textbf{97.93} $\pm$ 1.32 \\ 
    & WNTF & \textbf{93.07} $\pm$ 3.43 & 79.79 $\pm$ 27.16 & 97.62 $\pm$ 1.21 \\ 
    & BA & 92.33 $\pm$ 3.8 & 75.56 $\pm$ 32.58 & 97.47 $\pm$ 1.23 \\ 
    \hline

    \multirow{4}{*}{\rotatebox{90}{IMDB}} 
    & TB & 96.92 $\pm$ 1.98 & \textbf{98.38} $\pm$ 3.02 & 98.2 $\pm$ 1.37 \\ 
    & TF & 97.47 $\pm$ 1.79 & 87.67 $\pm$ 23.3 & \textbf{98.93} $\pm$ 1.11 \\ 
    & WNTF & \textbf{97.86} $\pm$ 1.49 & 94.55 $\pm$ 13.66 & 98.72 $\pm$ 1.2 \\ 
    & BA & 97.53 $\pm$ 2.24 & 75.25 $\pm$ 38.36 & 98.61 $\pm$ 1.27 \\ 
     \hline

    \multirow{4}{*}{\rotatebox{90}{Wiki\_PL}} 
    & TB & 95.82 $\pm$ 4.7 & \textbf{96.44} $\pm$ 6.6 & 97.69 $\pm$ 1.3 \\ 
    & TF & 96.36 $\pm$ 3.56 & 72.22 $\pm$ 25.38 & 97.52 $\pm$ 1.43 \\ 
    & WNTF & 96.36 $\pm$ 3.56 & 72.19 $\pm$ 25.42 & 97.52 $\pm$ 1.43 \\ 
    & BA & \textbf{96.72} $\pm$ 4.12 & 93.73 $\pm$ 10.0 & \textbf{98.35} $\pm$ 1.2 \\ 
    \hline

    \multirow{4}{*}{\rotatebox{90}{MultiEmo}} 
    & TB & 94.23 $\pm$ 2.9 & \textbf{98.03} $\pm$ 1.55 & 98.01 $\pm$ 1.28 \\  
    & TF & 95.41 $\pm$ 2.46 & 88.20 $\pm$ 15.36 & 98.25 $\pm$ 1.35 \\ 
    & WNTF & \textbf{96.50} $\pm$ 2.35 & 84.72 $\pm$ 19.81 & 98.64 $\pm$ 1.15 \\ 
    & BA & 95.9 $\pm$ 2.59 & 94.18 $\pm$ 11.58 & \textbf{98.79} $\pm$ 1.21 \\ 
    \hline

    \multirow{4}{*}{\rotatebox{90}{CSFD}} 
    & TB & 86.97 $\pm$ 11.50 & \textbf{92.37} $\pm$ 15.13 & 97.53 $\pm$ 1.28 \\ 
    & TF & 90.12 $\pm$ 7.21 & 68.73 $\pm$ 29.26 & 97.66 $\pm$ 1.40 \\ 
    & WNTF & \textbf{96.27} $\pm$ 3.64 & 77.35 $\pm$ 29.16 & \textbf{98.20} $\pm$ 1.48 \\
    & BA & 91.39 $\pm$ 6.23 & 84.23 $\pm$ 17.23 & 97.25 $\pm$ 1.30 \\ 
    \hline
    
    \multirow{4}{*}{\rotatebox{90}{CTDC}} 
    & TB & 97.69 $\pm$ 3.45 & \textbf{96.94} $\pm$ 5.97 & 98.25 $\pm$ 1.15 \\ 
    & TF & 97.69 $\pm$ 2.61 & 81.77 $\pm$ 22.2 & 98.14 $\pm$ 1.36 \\ 
    & WNTF & \textbf{99.56} $\pm$ 0.41 & 95.04 $\pm$ 9.74 & \textbf{99.1} $\pm$ 1.22 \\
    & BA & 98.32 $\pm$ 1.86 & 92.86 $\pm$ 8.84 & 97.92 $\pm$ 1.32 \\
    \hline

    \end{tabular}
\end{table}

\begin{figure}[htb!]
\centering
\includegraphics[width=0.92\linewidth]{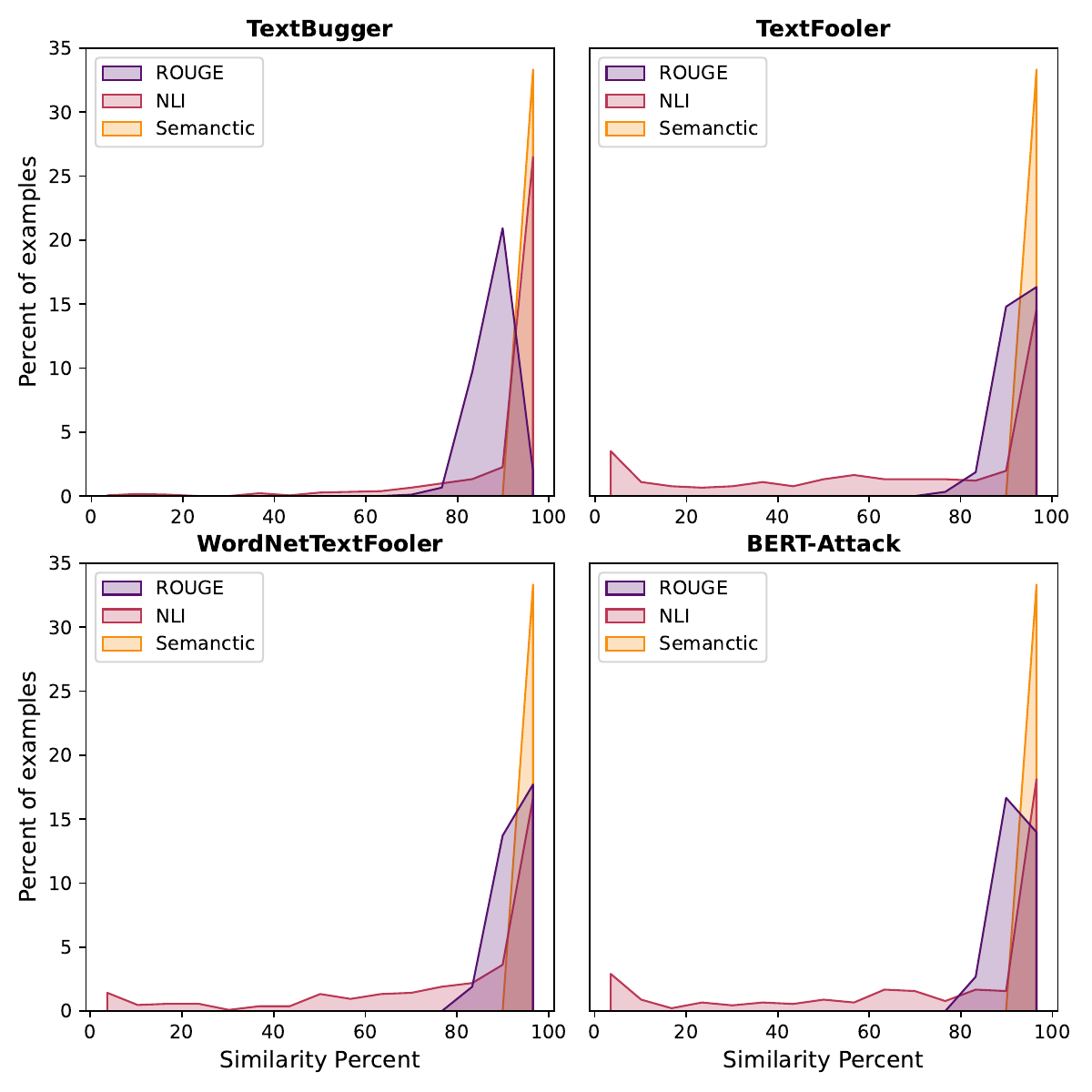}
\caption{Distribution of adversarial examples based on similarity to the original data for four evaluation metrics for the AG\_News dataset.} \label{chart:sims_ag_news}
\end{figure}

\begin{figure}[htb!]
\centering
\includegraphics[width=0.92\linewidth]{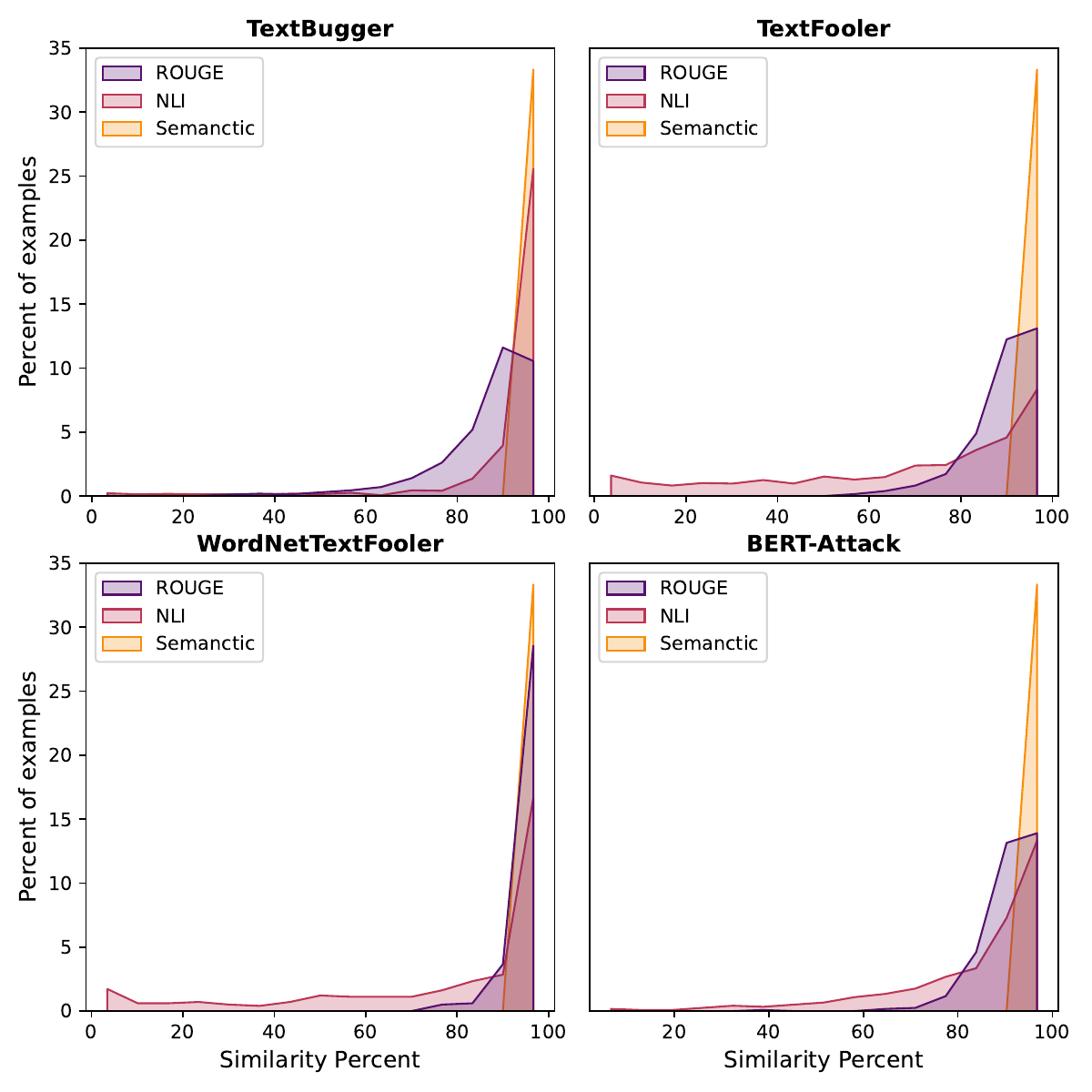}
\caption{Distribution of adversarial examples based on similarity to the original data for four evaluation metrics for the CSFD dataset.} \label{chart:sim_csfd}
\end{figure}

To further analyse the results of the obtained similarity metric, the similarity distribution of adversarial examples was examined, classified by attack methods, for two datasets, AG\_News and CSFD, which exhibited a high standard deviation in the NLI metric. The described distributions are presented in Figures \ref{chart:sims_ag_news} and \ref{chart:sim_csfd}. For the AG\_News the similarity distribution is long-tailed for adversarial examples generated by the TextFooler, WordNetTextFooler, and BERT-Attack methods. This correlates with the high standard deviation observed for these methods. In the CSFD dataset, a high standard deviation is observed for attacks using TextFooler and WordNetTextFooler. The reason for this phenomenon is evident in their distribution, which resembles the distribution observed for AG\_News. \textbf{In summary, the high standard deviation observed in similarity measurements using the NLI metric is caused by a subset of examples with very low similarity scores when evaluated with this method}.

\section{Mechanistic Interpretability of BERT-based classifiers}

\begin{figure}[htb]
\centering
\includegraphics[width=\linewidth]{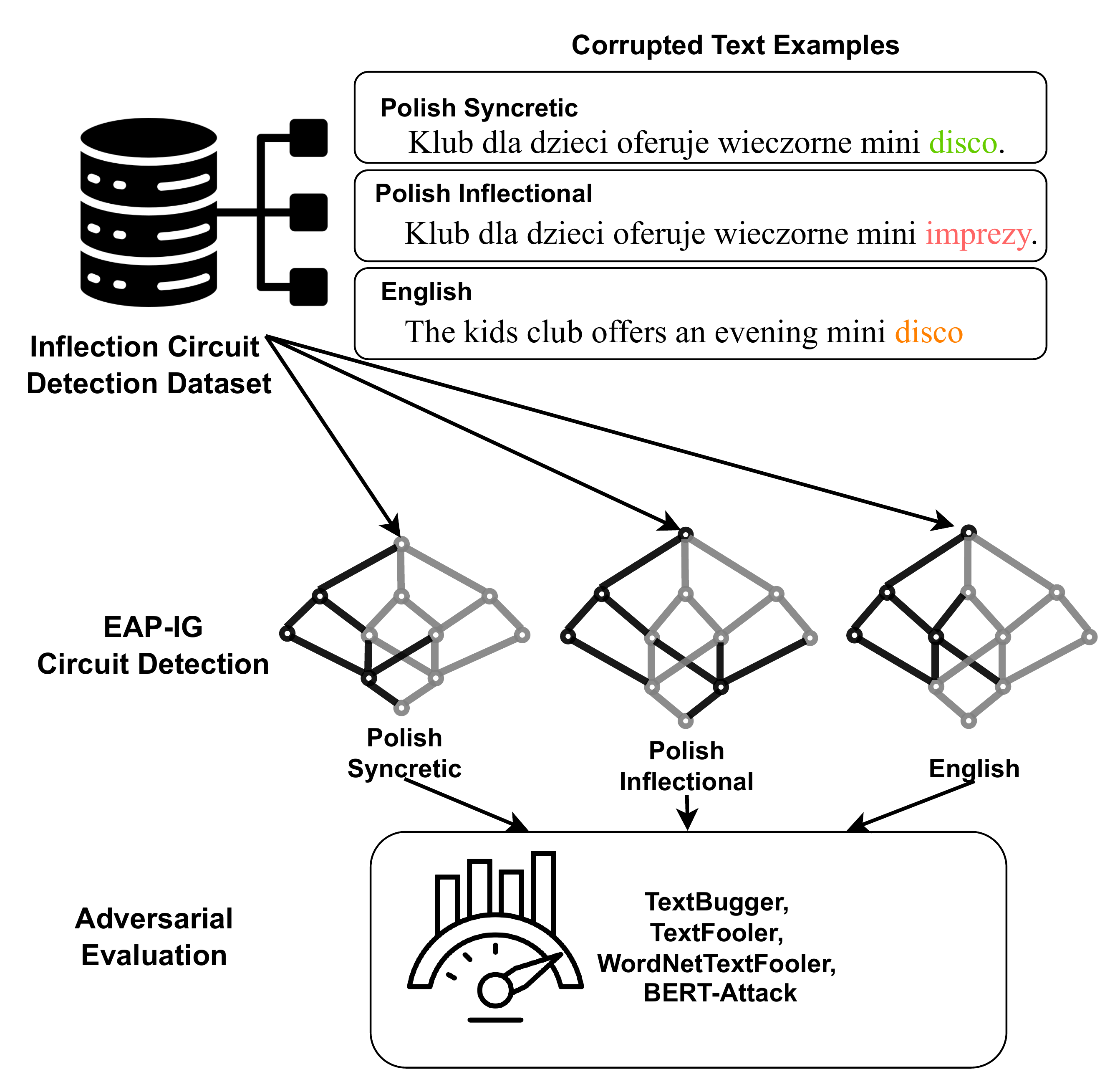}
\caption{Inflection circuit detection methodology with adversarial evaluation of identified circuits.} \label{mi}
\end{figure}

In order to interpret results obtained for inflectional and non-inflectional models, we crafted an \textbf{Inflectional Circuit Detection Dataset} based on  MutliEmo Corpus~\cite{multiemo}. Firstly, to fairly assess the performance across different languages, we trained the XLM-RoBERTa~\cite{DBLP:journals/corr/abs-1911-02116} language model on the Polish and English parts of the MultiEmo sentiment analysis task. The training part consist of 334 960 examples and test part of 41 710 examples. The trained model achieved an accuracy of 94\% on the test set. This part of the experimental analysis is illustrated in Figure~\ref{mi}.

\subsection{Inflectional Circuit Detection Dataset}\label{icdd}




Circuit discovery in transformer-based models typically relies on task-specific datasets. Methods such as EAP-IG are designed to assign importance scores to the edges between model components, effectively treating the model as a graph of interconnected elements. Constructing such circuits requires parallel datasets comprising clean (original) text and corresponding corrupted variants. These corrupted examples are crafted to introduce minimal perturbations to the original inputs, in a manner that isolates and emphasizes the model behaviour relevant to the targeted aspect. To enable the identification of circuits associated with inflectional processing, we constructed the Inflectional Circuit Detection Dataset, in which corrupted examples were specifically designed to reflect inflectional changes by altering a minimal number of tokens while preserving the overall semantic content. The dataset was crafted to follow the sentiment analysis task of MultiEmo~\cite{multiemo}. The first step was to obtain syncretic nouns (one form of the word represents multiple grammatical cases) using the Polish Grammatical Dictionary~\cite{sgjp}. Next we collected synonyms (words in the same synset) that have at least two syncretic forms and one inflexional, for that we used WordNet~\cite{wordnet} and Polimorf dataset~\cite{polimorf}. The vocabulary analysis presented above formed the basis for creating a parallel corpus. Sentences containing nouns that met the aforementioned conditions were selected from testset of the MultiEmo Corpus. These conditions entailed that the nouns had a minimum of two synonyms, one of which was syncretic and the other regular inflectional. Consequently, for each individual case, it is possible to create three parallel sentences differing only in the synonym used in each. Syncretic nouns are distinguished by their inability to undergo form changes in response to contextual variation, a property that contrasts with the capacity of regularly inflected nouns to demonstrate such changes. During the creation of the parallel corpus, the original noun was replaced by a synonym that is a syncretic noun and a synonym that has regular inflection. In the first case, the parallel corrupted sentence differs from the original only on the lexical level. In the second case, however, it may differ not only on the lexical level but also on the inflectional/formal level. Additional examples for the final version of the dataset were obtained with triplets of words, based on MultiEmo styled text,
then using Llama-3.3-70B-Instruct\footnote{\url{https://huggingface.co/meta-llama/Llama-3.3-70B-Instruct}} we expanded the number of examples, so that each text appears 10 times in the dataset with different words substituted around clean-corrupted pairs. In the final version of the dataset we excluded example pairs for which the model's predictions remain unchanged between the clean and corrupted runs. The final dataset consists of \textbf{3 dataset versions each with 244 examples}: 
\begin{itemize}
    \item \textbf{Polish Syncretic} (syncr.) containing: Polish Clean Text (Original Text) and Polish Corrupted Text with word replaced to its syncretic synonym,
    \item \textbf{Polish Inflectional} (inflex.) containing: Polish Clean Text (Original Text) and Polish Corrupted Text with word replaced to its inflexional synonym,
    \item \textbf{English} (EN) containing: English Clean Text (Original Text) and English Corrupted Text with word replaced to its synonym.
\end{itemize}



\subsection{Edge Attribution Patching}
In recently growing field of \textbf{Mechanical Interpretability (MI)} researchers aim to explain the behaviour of the model by measuring changes inside the model. One of the techniques used in MI is circuit discovery. Identifying circuits in language models can be expensive due to the challenge of pinpointing essential components and the main connections linking them. In order to effectively obtain circuits \citep{nanda2023} developed \textbf{Edge Attribution Patching (EAP)}, a technique that estimates the impact of connections with minimal computational passes through the model. The \textbf{Edge Attribution Patching with Integrated Gradients (EAP-IG)} \cite{hanna2024have} is an improvement over the EAP method which includes the integrated gradient technique (IG) \cite{DBLP:conf/icml/SundararajanTY17} into the circuit recognition. The integrated gradients addition aims to improve the EAP method when facing zero gradients regions between clean and corrupted response.

\subsection{Circuits Detection}
The EAP-IG~\cite{hanna2024have} circuit detection method represents the transformer model as a directed graph of connected components (attention heads and MLP layers). The method employs a predefined task-specific metric to assign a score to each edge of the graph. In our case for the MultiEmo task, we defined metric as the average predictive confidence of the correct class (negative or positive sentiment). Next the method select components connected with N highest rated edges, were N is a parameter. Edges that are more critical to task performance are assigned higher scores than those of lesser importance. Following this logic, it is possible to extract circuits (subgraphs of the model’s component graph) that can perform the task using a minimal number of components. Using the EAP-IG method and the Inflection Circuit Detection Dataset (in three versions), we calculated MultiEmo sentiment analysis circuits in 6 sizes. The variants included constructing circuits with: 50, 75, 100, 150, 200 and 300 model graph edges with highest score. \textbf{In summary, we constructed six circuit variants for Polish Syncretic, Polish Inflectional and English version of the Inflection Circuit Detection Dataset}. The performance of the circuits obtained on the circuit detection dataset is presented in Table~\ref{table:graph_baseline}. It should be noted that increasing the number of edges does not always lead to improved circuit performance; therefore, it is essential to evaluate circuits with varying number of edges. The instability in the metric values observed with an increasing number of edges number can be attributed to the methodology of the measurement technique. Circuit performance is evaluated by running the model on clean texts while applying corrupted activations to all edges not included in the circuit. This process is performed for each sample in the dataset. Since different samples may rely on different subsets of edges, a circuit represents an averaged optimal configuration. Consequently, including additional edges in the circuit may improve performance for some examples while simultaneously degrading it for others, leading to the observed instability in performance as the circuit size increases.


\begin{table}[htb]
    \caption{Task performance of the EAP-IG-generated circuits across six different sizes, alongside the baseline model performance. The baseline means run without applying the corrupted activations.}
    \label{table:graph_baseline}
    \centering
    \begin{tabular}{|l|rrr|}
    \hline
    \textbf{Graph size} & \textbf{syncr.} & \textbf{inflex.} & \textbf{EN}\\
    \hline
    Baseline   & 93.61       & 93.55  & 89.92 \\ \hline
    50         & \textbf{70.16}       & \textbf{89.11}       & \textbf{85.48} \\
    75         & 58.06       & 45.16       & 59.68 \\
    100        & 58.06       & 52.02       & 46.77 \\
    150        & 53.63       & 44.35       & 52.82 \\ 
    200        & 48.79       & 45.97       & 50.40 \\
    300        & 55.24       & 51.21       & 54.84 \\ \hline
    \end{tabular}
\end{table}

\begin{figure}[htb]
\centering
\includegraphics[width=1\linewidth]{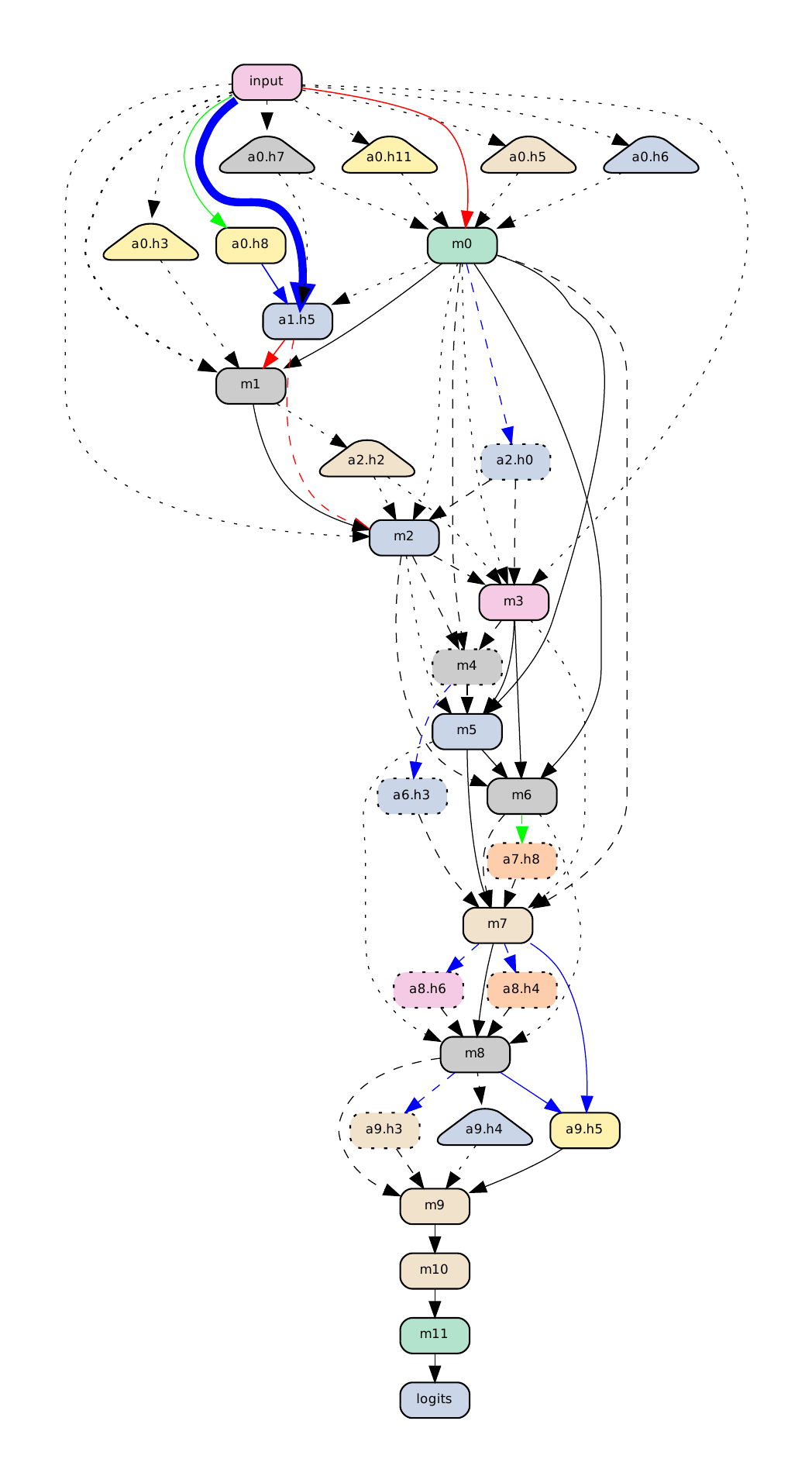}
\caption{Graph comparison of Polish syncretic vs. Polish inflectional circuits, showing the 50 most important edges. Nodes represent model submodules: rectangular (present in syncretic), dotted (removed in inflectional), and triangular (added in inflexional). Edge thickness reflects connection importance.} \label{graph}
\end{figure}


Circuits can be visualized for interpretability; a comparison of the Polish Syncretic and Polish Inflectional circuits in size of 50 top edges is presented in Figure~\ref{graph}. The input corresponds to the text embedding layer, whereas the logits represent the output layer logits produced by the model. The components shown in Figure~\ref{graph} represent MLP layers and attention heads, where labels such as m1 denote MLP layer 1 and a0.h1 indicates the first attention head in layer 0. The comparison of circuits is visualized by overlaying the Polish Inflectional graph onto the Polish Syncretic graph. Rectangular nodes indicate components present in the Syncretic circuit; if a node has a dotted border, it was not part of the Inflectional circuit. Similarly, edges drawn with dashed lines were present in the Syncretic circuit but absent from the Inflectional one. Conversely, components introduced in the Inflectional circuit are represented by triangular nodes and dotted edges. 

The graph comparison clearly reveals that the Inflectional circuit consistently introduces a series of attention heads in Layer 0 specifically, heads a0.h3, a0.h5, a0.h6, a0.h7, and a0.h11. This pattern persists across different circuit sizes, suggesting that these heads may be characterized as inflection-specific, as they are selectively activated in response to inflectional changes in the corrupted texts.




\subsection{Circuits Robustness against Adversarial Examples}

In this section, we utilized the three identified circuit variants: Polish Syncretic, Polish Inflectional and English to investigate whether the presence of inflection influences model robustness against adversarial examples. Adversarial examples were generated using the MultiEmo dataset and four established attack methods: BERT-Attack, TextBugger, TextFooler, and WordNetTextFooler. These examples were then used to assess the robustness of each circuit variant. The results of these experiments are presented in Table~\ref{table:graph_eval}. 
\begin{table}[htb!]
    \caption{Comparison of EAP-IG generated circuits (Polish syncretic, Polish inflectional and English) in prediction under adversarial examples generated with four methods: BERT-Attack, TextBugger, TextFooler and WordNetTextFooler, divided by circuit size.}
    \label{table:graph_eval}
    \centering
    \begin{tabular}{|l|rrr|rrr|}
    \hline
    \multirow{2}{*}{\shortstack{\textbf{Graph} \\ \textbf{size}}}
     & \multicolumn{3}{|c|}{\textbf{BERT-Attack}} & \multicolumn{3}{|c|}{\textbf{TextBugger}} \\
     & syncr. & inflex. & EN & syncr. & inflex. & EN \\
    \hline
    50 & 67.19 & \textbf{87.50} & 84.38 & 56.01 & \textbf{85.71} & 76.95 \\
    75 & \textbf{50.00} & 48.44 & 45.31 & 52.11 & \textbf{54.55} & 53.25 \\
    100 & 50.00 & \textbf{64.06} & 45.31 & \textbf{51.62} & 46.27 & 50.81 \\
    150 & \textbf{56.25} & 48.44 & 51.56 & \textbf{48.05} & \textbf{48.05} & 46.10\\
    200 & 43.75 & \textbf{54.69} & 39.06 & 50.49 & \textbf{51.79} & 46.92\\
    300 & \textbf{57.81} & 48.44 & 46.88 & \textbf{55.03} & 51.79 & 50.00\\ \hline
    Mean & 54.17 & \textbf{58.59} & 52.08 & 52.22 & \textbf{56.36} & 54.00 \\
    \hline
    \multirow{2}{*}{\shortstack{\textbf{Graph} \\ \textbf{size}}} 
    & \multicolumn{3}{|c|}{\textbf{TextFooler}} & \multicolumn{3}{|c|}{\textbf{WordNetTextFooler}} \\  
    & syncr. & inflex. & EN & syncr. & inflex. & EN \\ \hline
    50 & 55.00 & \textbf{67.50} & 66.67 & 62.10 & \textbf{81.94} & 73.41 \\
    75 & 48.33 & \textbf{53.83} & 49.83 & 43.65 & \textbf{59.13} & 53.77 \\
    100 & 44.67 & \textbf{51.50} & 42.67 & 48.81 & \textbf{50.79} & 43.65 \\
    150 & 47.67 & \textbf{51.83} & 50.83 & 46.43 & \textbf{59.33} & 47.82 \\
    200 & 44.33 & 48.17 & \textbf{59.50} & \textbf{50.79} & 49.60 & 45.24 \\
    300 & 62.00 & \textbf{56.17} & 55.67 & 45.24 & 53.17 & \textbf{56.35} \\ \hline
    Mean & 50.33 & \textbf{54.83} & 54.20 & 49.04 & \textbf{58.53} & 54.99 \\
    \hline

    \end{tabular}
\end{table}

Inflection circuits outperformed the other variants in most of the evaluated scenarios. Although the Syncretic circuits showed slightly better performance at certain circuit sizes, the inflectional circuits consistently achieved the highest average robustness across all graph sizes and attack methods. These results align with expectations, given that the attack methods employed typically involve modifying parts or entire words through various perturbation algorithms. Considering that inflection refers to the morphological variation of words such as the use of suffixes and prefixes to convey grammatical relationships, it is reasonable to conclude that models trained to better understand inflectional structures are more resilient to adversarial attacks that manipulate internal word components.

\section{Discussion}

\subsection{Conclusion}
In this work, we evaluated how adversarial examples influence the robustness of the model in various language models.
Incorporating inflexion in language model training corpus can increase its robustness against adversarial examples which focuses on bugging words (e.g. changing some of the letters). 


Among the set of evaluated adversarial example generation methods, TextBugger demonstrates the highest effectiveness, as measured by both the attack success rate and the corresponding decrease in model accuracy. This observation suggests that perturbation strategies based on character-level modifications, such as inserting, deleting, or replacing individual letters, may be more disruptive to language models than approaches that rely on substituting entire words, as employed in methods like TextFooler and WordNetTextFooler. 


On the other hand, our analysis of adversarial example similarity levels indicated that TextBugger consistently produced statistically lower-quality attacks compared to the other evaluated methods. Among the similarity measurement techniques evaluated, Semantic Similarity and ROUGE demonstrated high stability. In contrast, the Natural Language Inference (NLI) method showed a high standard deviation in similarity scores across the dataset. Further investigation of the NLI results revealed that the method assigned scores close to zero for a subset of adversarial examples.



Regarding the mechanistic interpretability of adversarial examples in circuits with and without inflection, the research carried out demonstrated that, in the case of Polish, specific components within task circuits could be identified as responsible for processing the inflectional elements of the language. Comparative analysis between circuits with and without explicit inflection handling elements showed that the inclusion of inflection-specific components contributed to improved model robustness when subjected to adversarial perturbations. These findings suggest that accommodating morphological complexity at the circuit level may play a crucial role in enhancing the resilience of the model. Importantly, such inflectional components were not observed in the corresponding reference circuit for English, probably due to the relatively low morphological variability of the English language.

\subsection{Limitations}
One of the key limitations in researching less widely studied languages, such as Czech, is the limited availability of high-quality annotated datasets. This constraint is reflected in the performance of language models trained on Czech datasets such as CTDC and CSFD, where accuracy reaches respectively 83\% and 86\%. In contrast, models trained on Polish and English datasets consistently achieve accuracies exceeding 94\%, highlighting a significant disparity.

Another significant limitation is the high cost associated with producing a parallel dataset for circuit discovery. Accurate tokenization plays a crucial role in this process, as clean and corrupted samples must maintain an identical number of tokens to enable circuit computation. Constructing such datasets is therefore time-consuming and labor-intensive, often requiring substantial linguistic expertise and manual intervention to ensure alignment and consistency.

\subsection{Future Works}

In future work, our aim is to investigate additional dimensions of model robustness, with particular attention to the structural and functional properties of internal model circuits. Assessing the resilience of these circuits to adversarial perturbations represents a promising direction for understanding not only how models process linguistic features but also where their vulnerabilities lie. Among the most significant challenges in this line of research is the creation of high-quality datasets specifically tailored for circuit discovery. These datasets must be carefully designed to highlight the targeted linguistic phenomena while preserving token-level alignment between original and perturbed inputs, a requirement that demands considerable linguistic expertise, manual effort, and computational resources.





\begin{ack}

Financed by: 
(1) CLARIN ERIC (2024–2026), funded by the Polish Minister of Science (agreement no. 2024/WK/01);
(2) CLARIN-PL, the European Regional Development Fund, FENG program (FENG.02.04-IP.040004/24);
(3) statutory funds of the Department of Artificial Intelligence, Wroclaw Tech;
(4) the EU project 'DARIAH-PL', under investment A2.4.1 of the National Recovery and Resilience Plan.
(5) the European Regional Development Fund as part of the 2014-2020 Smart Growth Operational Program (POIR.04.02.00-00C002/19); and (6) by the National Science Center, Poland, grant number 2018/29/B/HS2/02919.

\end{ack}



\bibliography{mybibfile}

\end{document}